\documentclass[letterpaper]{article} 
\usepackage[submission]{aaai23}  
\usepackage{times}  
\usepackage{helvet}  
\usepackage{courier}  
\usepackage[hyphens]{url}  
\usepackage{graphicx} 
\usepackage{natbib}  
\usepackage{caption} 
\usepackage{algorithm}
\usepackage{algorithmic}
\usepackage{booktabs}       
\usepackage{amsfonts}       
\usepackage{nicefrac}       
\usepackage{microtype}      
\usepackage{xcolor}         
\usepackage{enumitem}
\usepackage{makecell}
\usepackage{amssymb}
\usepackage{amsmath}
\usepackage{amsthm}
\usepackage[switch]{lineno}
\usepackage{amsmath} 
\usepackage{subfigure}
\usepackage{multirow}

\usepackage{colortbl}
\newtheorem{theorem}{Theorem}
\newtheorem{definition}{Definition}

\definecolor{gray}{RGB}{222,222,222}
\allowdisplaybreaks
\usepackage{newfloat}
\usepackage{listings}
\DeclareCaptionStyle{ruled}{labelfont=normalfont,labelsep=colon,strut=off} 
\floatstyle{ruled}
\newfloat{listing}{tb}{lst}{}
\floatname{listing}{Listing}
\title{Domain Discrepancy Aware Distillation for Model Aggregation in Federated Learning}
\author{
    Shangchao Su,
    Bin Li,
    Xiangyang Xue \\
    Fudan University
}
\affiliations{
    \textsuperscript{\rm 1}Association for the Advancement of Artificial Intelligence\\

    1900 Embarcadero Road, Suite 101\\
    Palo Alto, California 94303-3310 USA\\
    publications23@aaai.org
}

\usepackage{bibentry}

\begin{document}
\maketitle
\begin{abstract}
Knowledge distillation has recently become popular as a method of model aggregation on the server for federated learning. It is generally assumed that there are abundant public unlabeled data on the server. However, in reality, there exists a domain discrepancy between the datasets of the server domain and a client domain, which limits the performance of knowledge distillation. How to improve the aggregation under such a domain discrepancy setting is still an open problem. In this paper, we first analyze the generalization bound of the aggregation model produced from knowledge distillation for the client domains, and then describe two challenges, server-to-client discrepancy and client-to-client discrepancy, brought to the aggregation model by the domain discrepancies. Following our analysis, we propose an adaptive knowledge aggregation algorithm FedD3A based on domain discrepancy aware distillation to lower the bound. FedD3A performs adaptive weighting at the sample level in each round of FL. For each sample in the server domain, only the client models of its similar domains will be selected for playing the teacher role. To achieve this, we show that the discrepancy between the server-side sample and the client domain can be approximately measured using a subspace projection matrix calculated on each client without accessing its raw data. The server can thus leverage the projection matrices from multiple clients to assign weights to the corresponding teacher models for each server-side sample. We validate FedD3A on two popular cross-domain datasets and show that it outperforms the compared competitors in both cross-silo and cross-device FL settings.

\end{abstract}

\section{Introduction}
\label{intro}
Federated Learning (FL)~\cite{mcmahan2017communication,yang2019federated} is proposed and achieves rapid development for the purpose of privacy protection. It allows multiple clients to conduct joint machine learning training in which the client datasets are not shared. In each round of communication, the server sends the global model (a random initialization model for the first round) to the clients. Then each client utilizes its local data to train a local model based on the global model, and the server gathers the updated local models of the clients for aggregation. The aggregated model will eventually be able to handle the data of multiple clients after going through several rounds of communication.

FedAvg~\cite{mcmahan2017communication} is a traditional FL algorithm that obtains the aggregated model by directly averaging the local model parameters on the server. However, parameter averaging based aggregation methods can only be applied when the local models have exactly the same structure, which limits the application scenarios of FL. 
In some scenarios, such as automatic driving, the server often has a large amount of unlabeled data. Therefore, some studies~\cite{guha2019one,lin2020ensemble,gong2021ensemble,sturluson2021fedrad} replace parameter averaging with knowledge distillation (KD). They propose to use the public unlabeled data on the server, and leverage the ensemble output of the local models as a teacher to guide student's (aggregated model) learning. This solution can achieve an aggregated model from heterogeneous local models, and also reduce the number of communication rounds required for federated learning, because the server's training process can accelerate the convergence of the aggregated model.

However, the existing work based on KD does not consider the impact of the discrepancy between different domains. As we will point out in Section~\ref{method}, when there is a discrepancy between the public unlabeled data available on the server and the data in the client domains, the performance of the student (aggregated) model will drop significantly. Some works~\cite{zhang2021practical,zhang2021fedzkt} employ the data-free distillation by restoring pseudo samples on the server from the local models that fit the original domain data distributions. To use the data-free method in real natural images, the running mean and variance of the batch normalization layers must be obtained. However, some studies~\cite{yin2020dreaming,yin2021see} show that the mean and variance may leak the user's training data. In addition, the data-free method requires a large amount of additional calculation cost. Therefore, in this paper, we focus on the setting that the server has public unlabeled data. Now the problem is: how can we extract as much knowledge as possible from available unlabeled data on the server to maximize the performance of the aggregated model obtained through distillation? 

To answer this question, we define a domain as a pair $(\mathcal{D}, f)$ consisting of a distribution $\mathcal{D}$ on the input data $\mathcal{X}$ and a labeling function $f$,  and then analyze the generalization error of the aggregated model obtained by KD in the client domains. We find that when the teacher models (local models) originate from different domains, there are two main factors that affect the performance of the aggregated model: 1) Server-to-Client (S2C) discrepancy. When the domain discrepancy between the server domain and each client domain increases, the performance of the aggregated model on the client side will decrease. 2) Client-to-Clients (C2C) discrepancy. There is a knowledge conflict among multiple client models. Since the teacher models are learned from different client domains, the sample in the server domain is likely to obtain different predictions from different teacher models, which may severely interfere with each other. Based on these two factors, we can see that if we can reasonably take into account the similarity between the server domain and different client domains, and assign the most appropriate teacher models to different samples in the server domain, the generalization error upper bound can be reduced.

Motivated by our analysis, we propose an adaptive knowledge aggregation method, Domain Discrepancy Aware Distillation, called FedD3A. During distillation, we assign independent teacher weights to a server-side sample based on how similar this sample is to the client domains. Then we can reduce the knowledge conflict between different teacher models. In each round, the client extracts the features of the local data using the backbone of the global model, and then calculates the subspace projection matrix of the local feature space. The server obtains the projection matrices from the clients, and calculates the angles between the server data and the local feature space to measure similarity without accessing client data features. Overall, our contributions are as follows:
\begin{itemize}
\item By analyzing the generalization error of the aggregated model in the client domains, we point out two possible reasons why distillation-based model aggregation performance drops when there is a discrepancy between the server unlabeled data and the client domain data, namely Server-to-Client (S2C) discrepancy and Client-to-Clients (C2C) discrepancy.

\item Motivated by our analysis, we propose an aggregation method FedD3A based on domain discrepancy aware distillation, which further exploits the potential of abundant unlabeled data on the server.

\item To validate our method, we conduct extensive experiments on several datasets. The results show that compared with baselines, our method has a significant improvement in both cross-silo and cross-device FL settings. 

\end{itemize}

\section{Related Work}
\textbf{Federated Learning.}
FedAvg~\cite{mcmahan2017communication} proposes the federated averaging method. In each round of communication, a group of clients is randomly selected, the initial model is sent to all clients for training, and then the models trained by the clients are collected by the server. The aggregated model is obtained by averaging the model parameters of the clients. Some works~\cite{li2019convergence,sahu2018convergence} demonstrate the convergence of FedAvg and point out that the performance of FedAvg will degrade when different client datasets are non-iid distributed. A lot of research has tried to solve the non-iid problem encountered by FedAvg. One type of work~\cite{conf/mlsys/LiSZSTS20,karimireddy2020scaffold,reddi2020adaptive,wang2020tackling,WangYSPK20,singh2020model,su2022one} attempts to improve the fitting ability of the global aggregated model. The other type of work~\cite{DBLP:conf/nips/DinhTN20,fallah2020personalized,hanzely2020lower,DBLP:conf/iclr/LiJZKD21} seeks to establish the personalized federated learning (pFL), in which the clients can train different models with different parameters.

\textbf{Knowledge Distillation.}
Knowledge distillation~\cite{hinton2015distilling} is a knowledge transfer approach and is initially proposed for model compression. Usually, a larger model is used as the teacher model, and the knowledge of the teacher model is transferred to the student model by letting the smaller student model learn the output of the teacher model. The techniques of KD are mainly divided into logits-based distillation~\cite{hinton2015distilling,li2017learning}, feature-based distillation~\cite{romero2014fitnets,huang2017like,yim2017gift}, and relation-based distillation~\cite{park2019relational,liu2019knowledge,tung2019similarity}. Some works~\cite{du2020agree,shen2019meal} have also conducted research on multi-teacher distillation. \cite{du2020agree} tries to make the gradient of the student model close to that of all teacher models by multi-objective optimization. \cite{shen2019meal} uses adversarial learning to force students to learn intermediate features similar to multiple teachers. There are some studies~\cite{yin2020dreaming,lopes2017data,chawla2021data} on data-free distillation, which use pseudo samples generated by the teacher model to replace real data.

\textbf{Federated Learning with Knowledge Distillation.}
There are several ways of applying KD in FL: 1) The first is to perform distillation on the clients~\cite{yao2021local,wu2022communication,zhu2021data}, treating the global aggregated model as the teacher. 2) In~\cite{gong2021ensemble,gong2022preserving,sun2020federated}, the server and all the clients share a public unlabeled dataset. The predictions of different models on the dataset are transmitted among all parties to perform distillation. This is often used for personalized federated learning. 
3) The third way is to directly use the client models as the teachers. \cite{lin2020ensemble,guha2019one} suppose the server has unlabeled data and use an ensemble of multiple client models as the teacher. The average output of the teacher model is then used to calculate the distillation loss. \cite{sturluson2021fedrad} proposes using the median-based scores instead of the average logits of teacher outputs for distillation. In addition, the date-free method has also been applied on the server, \cite{zhang2021practical,zhang2021fedzkt} try to learn a generative model based on the ensemble of client models. However, high-quality pseudo samples rely on the running mean and variance carried by BN layers in the client models, which may reveal privacy~\cite{yin2020dreaming,yin2021see}. The data-free method requires a large amount of computational cost, and there is no evidence that it can be applied to tasks other than image classification.
Nevertheless, data-free method is orthogonal to FedD3A, and all pseudo samples can be used as our training data.

\section{Proposed Method}
\label{method}
\subsection{Notations and Analysis}
\label{ana}
Following the domain adaptation~\cite{ben2010theory} field, during the analysis, we consider the binary classification task, where a \textit{domain} is defined as a pair consisting of a distribution $\mathcal{D}$ on the input data $\mathcal{X}$ and a labeling function $f: \mathcal{X} \to \{ 0,1 \}$. In FL, we have $K$ client domains, $(\mathcal{T}_{k}, f_{k}), k=1,\cdots,K$. The server domain is $(\mathcal{S}, f_{s})$, where $f_{s}$ is unknown because the server has only unlabeled data. 

Knowledge distillation on the server takes $K$ client models $h_k,k=1, \cdots, K$ as the teacher models, and uses the unlabeled data $\mathcal{S}$ to train the aggregated model $h$ (student). KD aims to minimize the generalization error of the aggregated model on multiple clients in each round:
\begin{linenomath}\begin{align}
\sum_k\lambda_k\mathrm{R}_{\mathcal{T}_k}(h)=\sum_k\lambda_k\underset{\mathbf{x} \sim \mathcal{T}_k}{\mathbb{E}}[|h(\mathbf{x})- f_{k}(\mathbf{x})|]
\end{align}\end{linenomath}
where $\lambda_k$ denotes the importance of the $k$-th client domain and defaults to $\frac{1}{K}$, $\mathrm{R}$ is the risk. When performing multi-teacher distillation, it can be viewed as adding pseudo-labels to $\mathcal{S}$. That is, we can view $f_s$ as $\hat{f_{s}}(\mathbf{x})=\text{sign}(\frac{1}{K}\sum_k h_k (\mathbf{x }))$. Next, we analyze the generalization error of the aggregated model in the client domains through the theory of \textit{domain adaptation}~\cite{ben2010theory}.

\begin{definition}[$L^1$-distance~\cite{ben2010theory}]
$\mathcal{B}$ is the set of measurable subsets under $\mathcal{D}_{1} $ and $ \mathcal{D}_{2}$. The $L^{1}$-distance between two distributions $\mathcal{D}_{1} $ and $ \mathcal{D}_{2}$ is defined as:
\begin{linenomath}\begin{align}
d_{1}\left(\mathcal{D}_{1}, \mathcal{D}_{2}\right)=2 \sup _{B \in \mathcal{B}}\left|\underset{\mathcal{D}_{1}}{\operatorname{Pr}}(B)-\underset{\mathcal{D}_{2}}{\operatorname{Pr}}(B)\right|
\end{align}\end{linenomath}
\end{definition}
Then we have:
\begin{theorem}
\label{t1}
Given the server data $\mathcal{S}$ , the client domains $(\mathcal{T}_{k}, f_{k}), k=1,\cdots,K$, and a hypothesis space $\mathcal{H}$, $\mathrm{R}_{\mathcal{S}}(h)=\underset{\mathbf{x} \sim \mathcal{S}}{\mathbb{E}} \ell(h(\mathbf{x}),\hat{f_{s}}(\mathbf{x}))$ . $\forall h\in\mathcal{H}$, the following holds:

\begin{linenomath}\begin{align}
\nonumber &\sum_k\lambda_k\mathrm{R}_{\mathcal{T}_k}(h) \leq \underbrace{\mathrm{R}_{\mathcal{S}}(h)}_{\text{\textcircled{1} Server risk}}+\underbrace{\sum_k\lambda_k d_{1}\left(\mathcal{S}, \mathcal{T}_k\right)}_{\text{\textcircled{2} S2C discrepancy}}+
\\ &
\underbrace{\min \left\{\sum_k\lambda_k\underset{\mathbf{x} \sim \mathcal{S}}{ \mathbb{E}}\mathcal{F}(\mathbf{x}), \sum_k\lambda_k\underset{\mathbf{x} \sim \mathcal{T}_k}{\mathbb{E}}\mathcal{F}(\mathbf{x})\right\}}_{\text{\textcircled{3} C2C discrepancy}}
\end{align}\end{linenomath}
where $\mathcal{F}(\mathbf{x})=|\hat{f_{s}}(\mathbf{x})-f_{k}(\mathbf{x})|$.
\end{theorem}
See Supplementary Materials for proof. Theorem~\ref{t1} implies that the error upper bound is related to three terms, from which we can summarize two major challenges for model aggregation through KD. The first challenge is related to the second term, that is server-to-client (S2C) discrepancy. When the difference between server domain data and client domain data is large, the risk upper bound will also increase. This domain discrepancy can naturally be reduced if we can access the client domain data and perform data augmentation through image mixing. Unfortunately, in the context of federated learning, we cannot obtain client data. The second challenge is related to the third term, we call it client-to-clients (C2C) discrepancy, that is, the difference between a single client and all clients. In the extreme case that we have only one client $(\mathcal{T}_1,f_1)$, the third term becomes:
\begin{linenomath}\begin{align}
\nonumber
\min \left\{\underset{\mathbf{x} \sim \mathcal{S}}{ \mathbb{E}}\left[\left|\hat{f_{s}}(\mathbf{x})-f_{1}(\mathbf{x})\right|\right], \underset{\mathbf{x} \sim \mathcal{T}_1}{\mathbb{E}}\left[\left|f_{1}(\mathbf{x})-\hat{f_{s}}(\mathbf{x})\right|\right]\right\}
\end{align}\end{linenomath}
where $\underset{\mathbf{x} \sim \mathcal{T}_1}{ \mathbb{E}}\left[\left|f_{1}(\mathbf{x})-\hat{f_{s}}(\mathbf{x})\right|\right]=\underset{\mathbf{x} \sim \mathcal{T}_1}{ \mathbb{E}}\left[\left|f_{1}(\mathbf{x})-h_{1}(\mathbf{x})\right|\right]=\mathrm{R}_{\mathcal{T}_1}(h_1)$, which is a fixed small value, because the client model $h_1$ has been trained on the client domain $\mathcal{T}_1$. However, when we have multiple client models of different domains, due to the internal conflict between teacher models, the difference between the label function $\hat{f_{s}}$ and each client-side label function $f_k$ will be accumulated.


Based on the above analysis, we propose to assign the most reasonable pseudo-label to each sample in $\mathcal{S}$, thereby reducing the difference between the server's labeling function and the labeling functions $\{f_k\}_k$ of other client domains. Suppose there are samples in $\mathcal{S}$ that are similar to some client domains, and assume that we have a domain discriminant function $\text{dom}(\cdot)$ by which all samples in $\mathcal{S}$ can be labeled with their closest domain identity. Then $\mathcal{S}$ can be viewed as $\mathcal{S}=\sum\lambda_k\mathcal{S}_k, k=1,\cdots,K$, the samples in $\mathcal{S}_k$ are similar in distribution to $\mathcal{T}_k$, so the hypothesis $h_k(x)$ learned in the $k$-th domain can be used to approximate its labeling function. Now Theorem~\ref{t1} can be modified as:
\begin{theorem}
\label{t2}
For $\forall h\in\mathcal{H}$:
\begin{linenomath}\begin{align}
 &\nonumber\sum_k\lambda_k\mathrm{R}_{\mathcal{T}_k}(h) \leq \mathrm{R}_{\mathcal{S}}(h)+\sum_k\lambda_k d_{1}\left(\mathcal{S}_k, \mathcal{T}_k\right)+
\\ &
\min \left\{\sum_k\lambda_k\underset{\mathbf{x} \sim \mathcal{S}_k}{ \mathbb{E}}\mathcal{F}_{k}(\mathbf{x}), \sum_k\lambda_k\underset{\mathbf{x} \sim \mathcal{T}_k}{\mathbb{E}}\mathcal{F}_{k}(\mathbf{x})\right\}
\end{align}\end{linenomath}
where $\mathcal{F}_{k}(\mathbf{x})=|\hat{f}_{s_k}(\mathbf{x})-f_{k}(\mathbf{x})|$.
\end{theorem}

In Theorem~\ref{t2}, $\hat{f}_{s_k}(\mathbf{x})$ is approximated by $h_k(\mathbf{x})$. Since $h_k(\mathbf{x})$ is a hypothesis learned in $\mathcal{T}_k$, $\mathbb{E}[| h_k(\mathbf{x})-f_{k}(\mathbf{x})| ]$ is always a small value. Therefore, $\mathbb{E}[|\hat{f}_{s_k}(\mathbf{x})-f_{k}(\mathbf{x})| ]$ will also remain at a small value.


\subsection{Domain Discrepancy Aware Distillation}
\label{D3A}

\begin{figure*}[tb]
\centering
\includegraphics[width=0.9\linewidth]{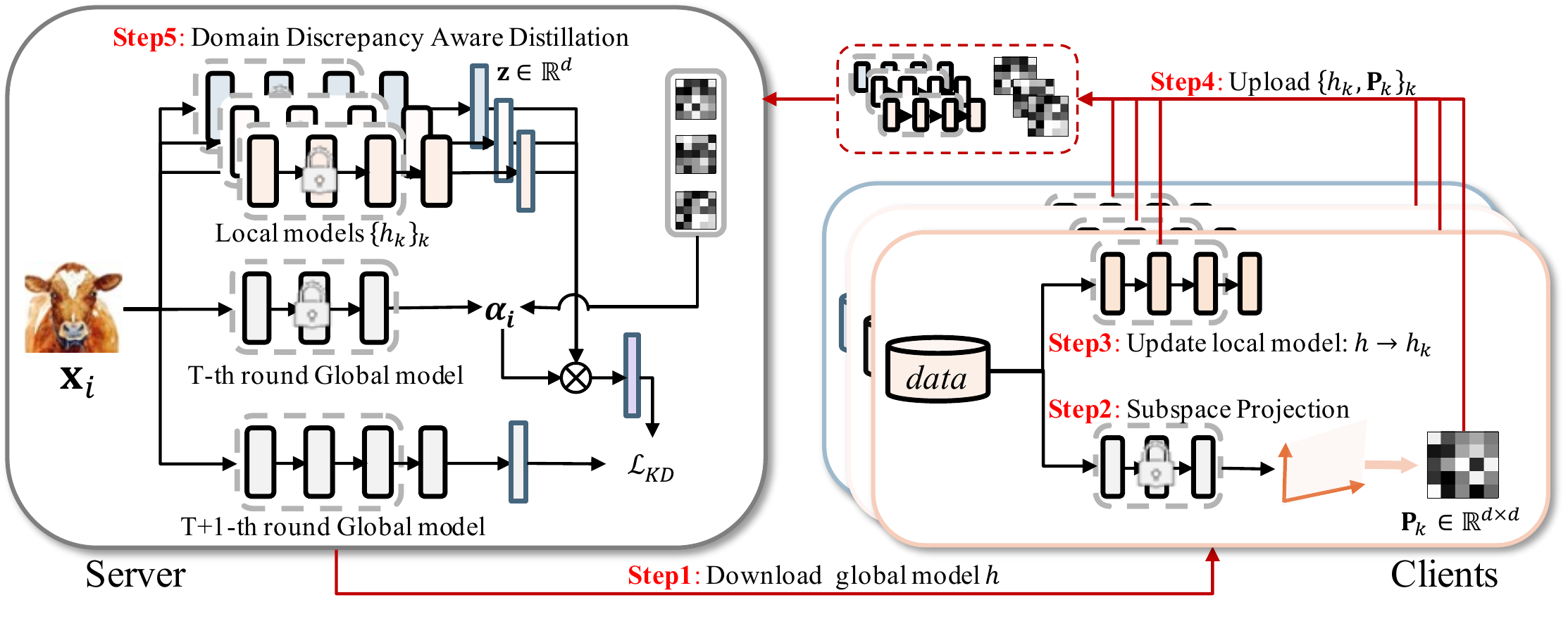}
\caption{Overview of FedD3A. In each round, the clients calculate the subspace projection matrices according to the features extracted from the local data by the current global model and then upload the model parameters and projection matrices. The server utilizes these projection matrices for adaptive teacher weighting.}
\label{framework}
\end{figure*}

From the above analysis, we establish an improved route, that is, use the domain discriminator $\text{dom}(\cdot)$ to perform teacher screening for individual samples on the server. Below we introduce the specific implementation method.

The ideal implementation is to obtain the data of all clients and directly calculate the data similarity, but it is obviously impossible in the FL scenario. One possible implementation of the domain discriminator is multi-party joint modeling. That is, one labels all data in $\mathcal{T}_k$ with a label $k$, and then $K$ clients jointly train a $K$-class  classifier by FL. This special setting has been studied in FedAws~\cite{yu2020federated}. FedAws performs multiple rounds of federated training to jointly learn a classifier when each client has only one class of data. However, if FedAws is adopted, an additional auxiliary FL task for domain classification has to be introduced before the main FL task, which greatly increases the communication cost and computational overhead. Therefore, below we propose a lightweight approach based on subspace projection to implement $\text{dom}(\cdot)$. The overview is illustrated in Figure~\ref{framework}. The overall framework consists of five steps. Steps 1 and 4 are the communication between the server and the clients. The other steps are as follows:

\textbf{Subspace Projection} (Step2).
Suppose that each client has dataset $\mathcal{D}=\left\{\left(\mathbf{x}_i, y_i\right)\right\}_{i=1}^{n}$, $n$ is the number of samples, for simplicity, we omit the client index. The feature matrix is $\mathbf{Z} \in \mathbb{R}^{n \times d}$, each row vector $\mathbf{z}_i \in \mathbb{R}^{d}$ in $\mathbf{\mathbf{Z}}$ is the feature of $\mathbf{x}_i$ extracted by backbone. Denote the subspace spaned by $\mathbf{Z}$ as $S$, then the projection matrix of $S$ is $\mathbf{P}=\mathbf{Z}^{\top}\left(\mathbf{Z} \mathbf{Z}^{\top}+ \alpha I\right)^{-1} \mathbf{Z}$. For any vector $\mathbf{u}$, $\mathbf{P}$ can project $\mathbf{u}$ onto space $S$, i.e., $\mathbf{P}\mathbf{u}$. The angle between the original vector $\mathbf{u}$ and the projection vector $\mathbf{P}\mathbf{u}$ reflects the proximity between $\mathbf{u}$ and $\mathbf{Z}$. We can use $\cos\left<\mathbf{P},\mathbf{P}\mathbf{u}\right>$ to measure the proximity. To avoid the matrix-inverse operation, we calculate $\mathbf{P}$ in an iterative manner~\cite{zeng2019continual}:
\begin{linenomath}\begin{align}
\hat{\mathbf{P}}_0 =\frac{\mathbf{I}}{\alpha} ,\quad \hat{\mathbf{P}}_i=\hat{\mathbf{P}}_{i-1}-\frac{\hat{\mathbf{P}}_{i-1}\overline{\mathbf{z}}_{i}\overline{\mathbf{z}}^\top_{i}\hat{\mathbf{P}}_{i-1}}{1+\overline{\mathbf{z}}^\top_{i}\hat{\mathbf{P}}_{i-1}\overline{\mathbf{z}}_{i}}
\label{proj}
\end{align}\end{linenomath}
where $\overline{\mathbf{z}}_{i}$ is the mean feature of the $i$-th batch.  After getting $\hat{\mathbf{P}}$, we have $\mathbf{P}=\mathbf{I}-\hat{\mathbf{P}}$. As shown in Figure~\ref{framework}, in each round, before local training, the client first extracts the features matrix $\mathbf{Z}$ of local data based on the global model, and then calculates the projection matrix $\mathbf{P}_i$ according to Eq.\ref{proj}. $\mathbf{P}_i$ will be uploaded to the server along with the model parameters. 

\begin{algorithm}[t]
	\begin{algorithmic}
	
		\STATE {{\bfseries Initialize:} 
		Initialize parameters $\boldsymbol{\theta}$ for the global model $h$, the number of global rounds $T_g$ and local epochs $T_l$
		}\\
		\FOR{$t=0,\dots,T_g-1$}
		\STATE Sample $m$ clients and send $\boldsymbol{\theta}$ to each selected client\\
		\FOR{$k=M_1,\dots,M_m$ in parallel}
		\STATE Calculate $\mathbf{P}_k$ using Eq.\ref{proj}\\
		\STATE Train $T_l$ epochs for each client: $\boldsymbol{\theta}_k\leftarrow \textit{LocalTrain}\left(h_k, \mathcal{T}_k\right)$\\
		\ENDFOR
		\STATE Initialize the student (global) model $\boldsymbol{\theta}=\textit{Average}(\boldsymbol{\theta}_{M_1},\cdots,\boldsymbol{\theta}_{M_m})$
		\FOR{each batch $\{\mathbf{x}_i\}$ in $\mathcal{S}$}
		\STATE Calculate $\{\boldsymbol{\hat{y}}_i\}$ using Eq.\ref{pseudo}
		\STATE Update the global model $h$ using Eq.\ref{LKD}.
		\ENDFOR
	\ENDFOR
	\end{algorithmic}
	\caption{FedD3A}\label{algalg}
\end{algorithm}

\textbf{Local Training} (Step3). After obtaining the projection matrix, the client uses the local dataset to train the local model. When the global model and the local model are homogeneous, the local model is initialized with the global model parameters. When the global model and the local model are heterogeneous, the global model and the local models have different structures. The client needs to use the global model as a teacher for training, that is, using the cross entropy loss and a distillation loss to guide the local model training at the same time. In the experimental part, we will show the effectiveness of FedD3A under both the homogeneous and heterogeneous setting.

\textbf{Server Training} (Step5). On the server, suppose the client indexes chosen in this round is $M_1,\cdots,M_m$, the backbone of global model is $\mathcal{B}$. For each sample, the corresponding teacher weights $\boldsymbol{\alpha}$ is calculated as follows:: 
\begin{linenomath}\begin{align}
\nonumber\mathbf{r} = \,\,& [\cos\left<\mathcal{B}(\mathbf{x}),\mathbf{P}_{M_1}\mathcal{B}(\mathbf{x})\right>, \cdots,\cos\left<\mathcal{B}(\mathbf{x}),\mathbf{P}_{M_m}\mathcal{B}(\mathbf{x})\right> ]\\
\label{alpha}
\boldsymbol{\alpha} = \,\,&\text{SoftMax}\left(\frac{\mathbf{r} -\text{mean}(\mathbf{r} )}{\sqrt{\text{var}(\mathbf{r} )}}\right)
\end{align}\end{linenomath}

The local models $h_k, k=M_1,\cdots,M_m$ are used as the teachers in distillation. Then for each sample $\mathbf{x}$ on the server, we have the pseudo-label:
\begin{linenomath}\begin{align}
\label{pseudo}
\boldsymbol{\hat{y}}= \sum_{i=1}^m {\alpha}_i h_{M_i}(\mathbf{x})
\end{align}\end{linenomath}
Now the loss function for distillation is:
\begin{linenomath}\begin{align}
\label{LKD}
\mathcal{L}_{KD} = D_{KL}(h(\mathbf{x})\|\boldsymbol{\hat{y}})
\end{align}\end{linenomath}
where $D_{KL}$ is the Kullback-Leibler divergence, $h$ can be initialized using the average parameters  of $h_{M_1},\cdots,h_{M_m}$. The overall method is described in Algorithm~\ref{algalg}.



\subsection{Discussions}
\textbf{Communication Cost.} Compared to FedAvg~\cite{mcmahan2017communication}, our method only needs to send the additional projection matrix to the server. Taking ResNet34 as an example, the number of parameters of the model is $21.8$M. Assuming that the data are $224\times224$ images and the feature extracted by the backbone is a $512$ dimensional vector. In this case, the number of parameters of the projection matrix is $512 * 512$, only accounting for $1.2\%$ ($512^2/21.8$M) of the total parameters of the model. If a technique such as SVD is used for matrix compression, this ratio can be further reduced. In addition, FedD3A can speed up the convergence and make the aggregated model have better performance, which makes the extra $1.2\%$ communication cost worthwhile.

\textbf{Privacy Protection.}
Our method only needs to exchange model parameters and projection matrices, without sharing the raw client-side datasets or features. Compared with the operation of directly transmitting the running mean images or the features extracted from the backbone in some works~\cite{peng2019federated,DBLP:conf/iclr/YoonSHY21}, the projection matrix can avoid directly further disclosing the user data features. It is also worth noting that, in the proposed method, it is not required to transmit the mean and variance in the BN layers of local models. Therefore the risk of restoring raw data~\cite{yin2020dreaming,yin2021see} can be avoided.

%
%

\textbf{Application to Heterogeneous Models.} 
Knowledge distillation based model aggregation methods only involve the outputs of the local models, thus allowing local models to have different structures. In this paper, we focus on the following heterogeneous model aggregation scenario: the clients have different model structures, the server aggregates these models into a public model structure. After the aggregated model is sent to the clients, each client uses KD for knowledge transfer. Ultimately, the server can obtain the global model while the clients retain their personalized models.

\section{Experiments}
In this section, to verify the ability of FedD3A, we first compare the overall performance of compared methods in cross-silo and cross-device settings. Then, we demonstrate the efficacy of FedD3A in heterogeneous model aggregation. Finally, we validate the effect of the proposed teacher model weighting strategy.

\begin{figure}[t]
    \centering
    \includegraphics[width=0.7\linewidth]{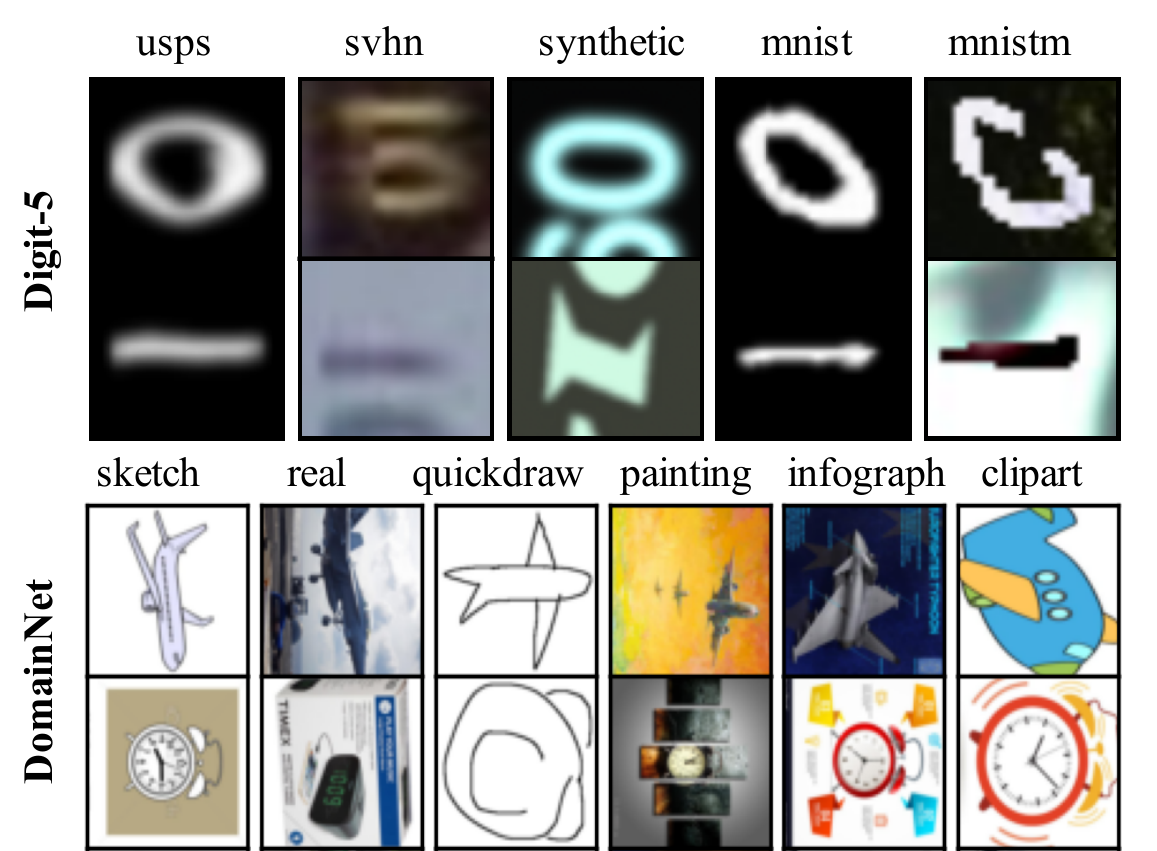}
    \caption{Digit-5 and DomainNet datasets.}
    \label{datasets}
\end{figure}

\textbf{Datasets and Models.}  For federated learning, we conduct experiments based on \emph{Digit-5}~\cite{zhao2020multi} which has five domains with 10 categories $32\times32$ images per domain, and \emph{DomainNet}~\cite{peng2019moment} which is a large-scale multi-domain dataset containing 6 domains with 345 categories $224\times224$ images per domain. The image examples in different domains are shown in Figure~\ref{datasets}. By default, ResNet50~\cite{he2016deep} is selected for DomainNet, ResNet18 is selected for Digit-5. All models are initialized using ImageNet pre-trained parameters.

\textbf{Baselines.} 
We take the following three different types of methods as our competitors: 1) \textit{FedAvg}~\cite{mcmahan2017communication}, a classical aggregation method based on parameter averaging. 2) \textit{FedProx}~\cite{li2020federated}, an improved method based on FedAvg for non-iid datasets. 3) \textit{FedDF}~\cite{lin2020ensemble}, a SOTA distillation based model aggregation method, where each teacher is treated equally.

\begin{figure*}[tb]
\centering
\subfigure[\textit{real} as Server]{
\includegraphics[width=0.24
\linewidth]{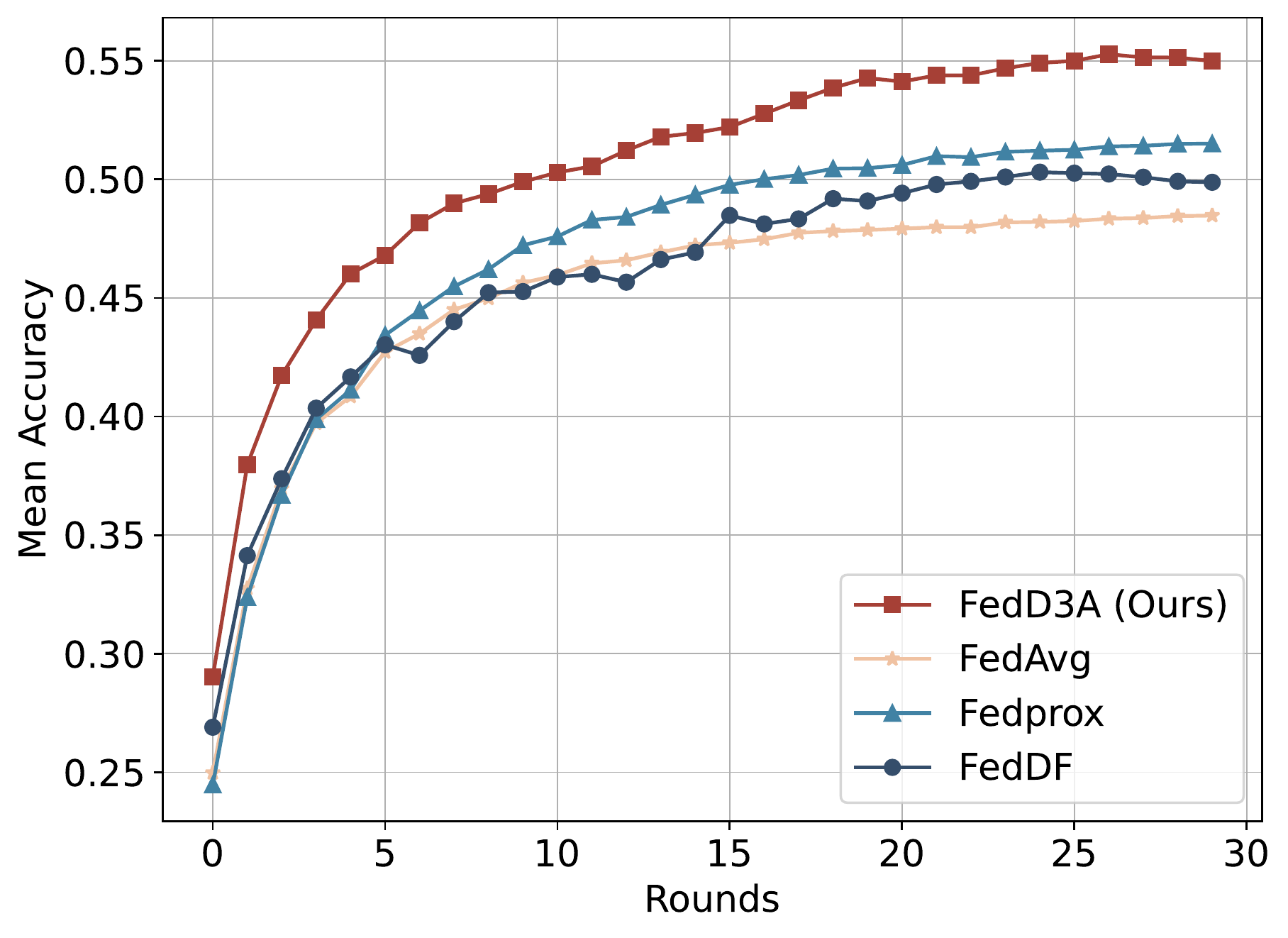}
}%
\subfigure[\textit{clipart} as Server]{
\includegraphics[width=0.24\linewidth]{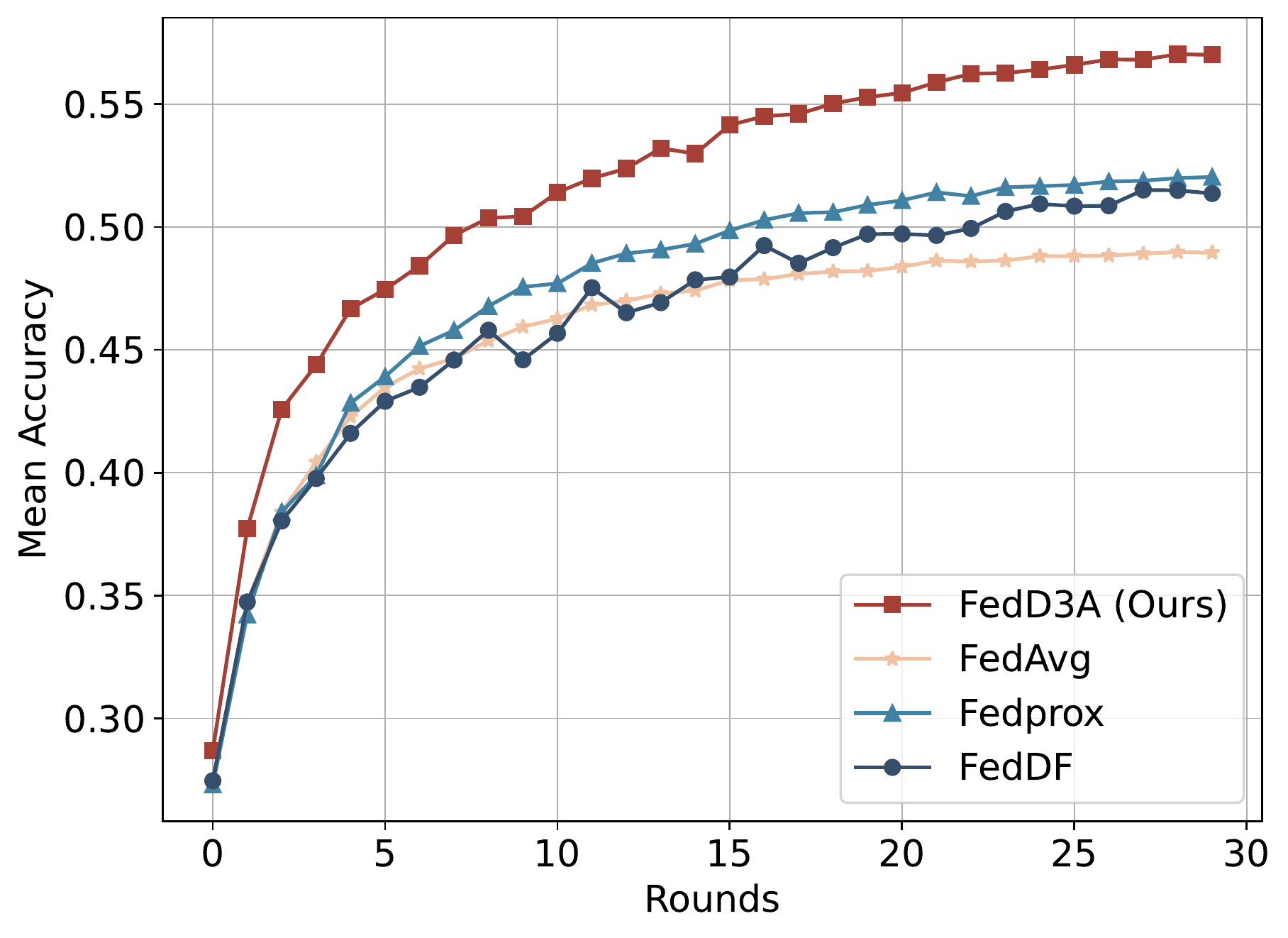}
}%
\subfigure[\textit{mnistm} as Server]{
\includegraphics[width=0.24
\linewidth]{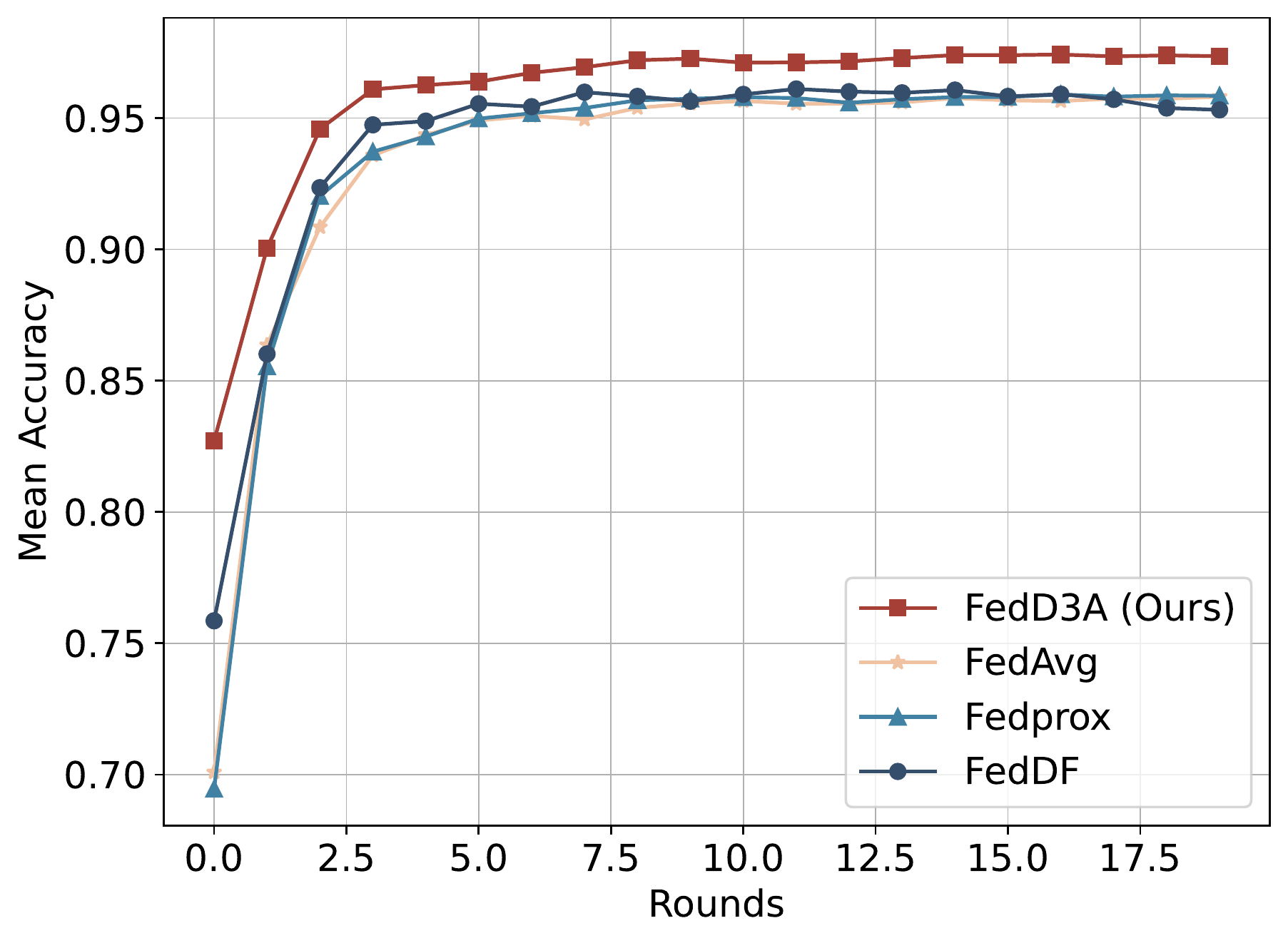}
}%
\subfigure[\textit{svhn} as Server]{
\includegraphics[width=0.24\linewidth]{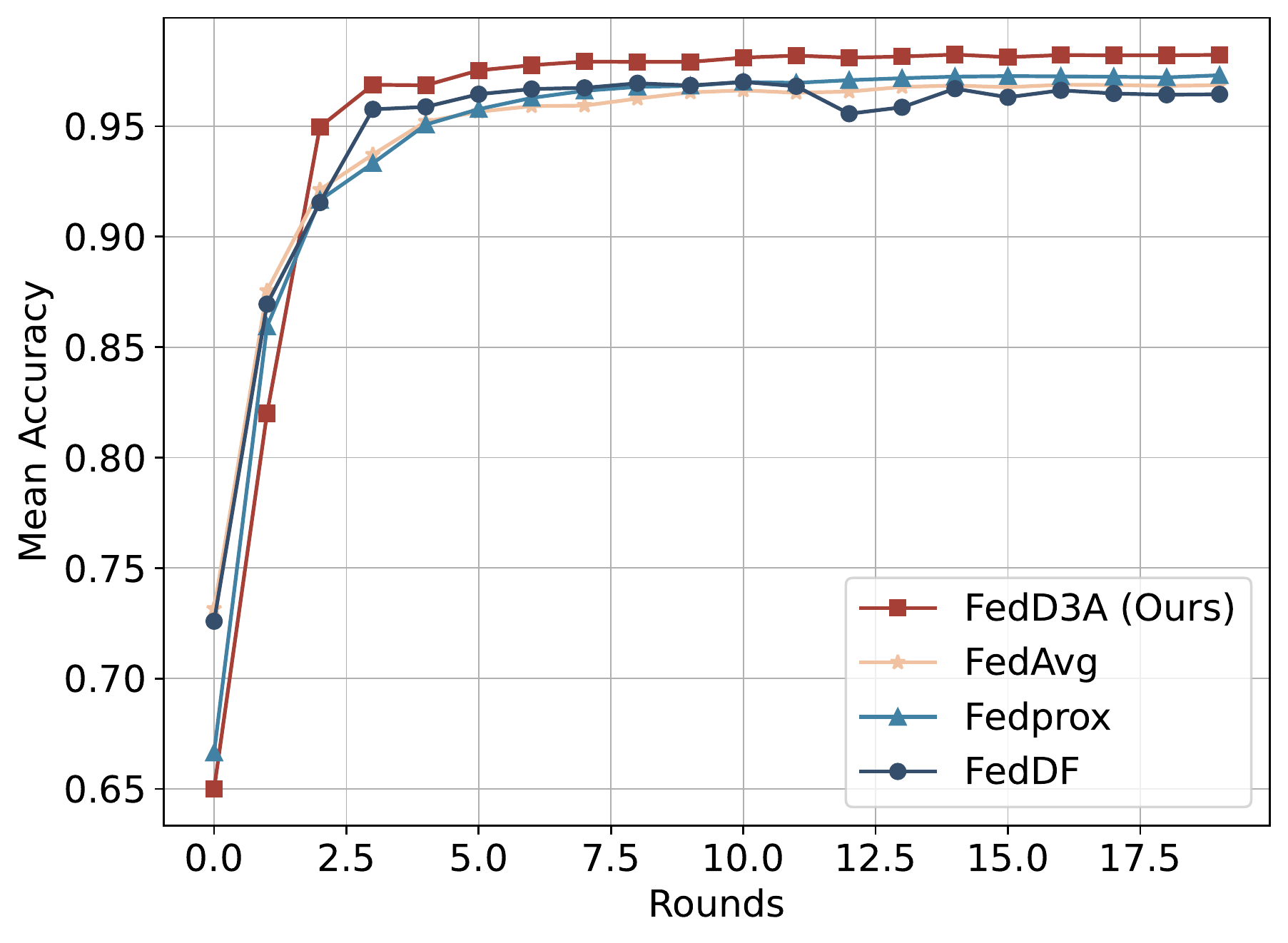}
}%
\caption{Average performance of the aggregated model on all clients in cross-silo setting. Ours method outperforms the baselines by a large margin. See the Supplementary Materials for the results about the other domains.}
\label{crosssilo}
\end{figure*}

\begin{figure*}[tb]
\centering
\subfigure[$\beta=0.2$]{
\includegraphics[width=0.24
\linewidth]{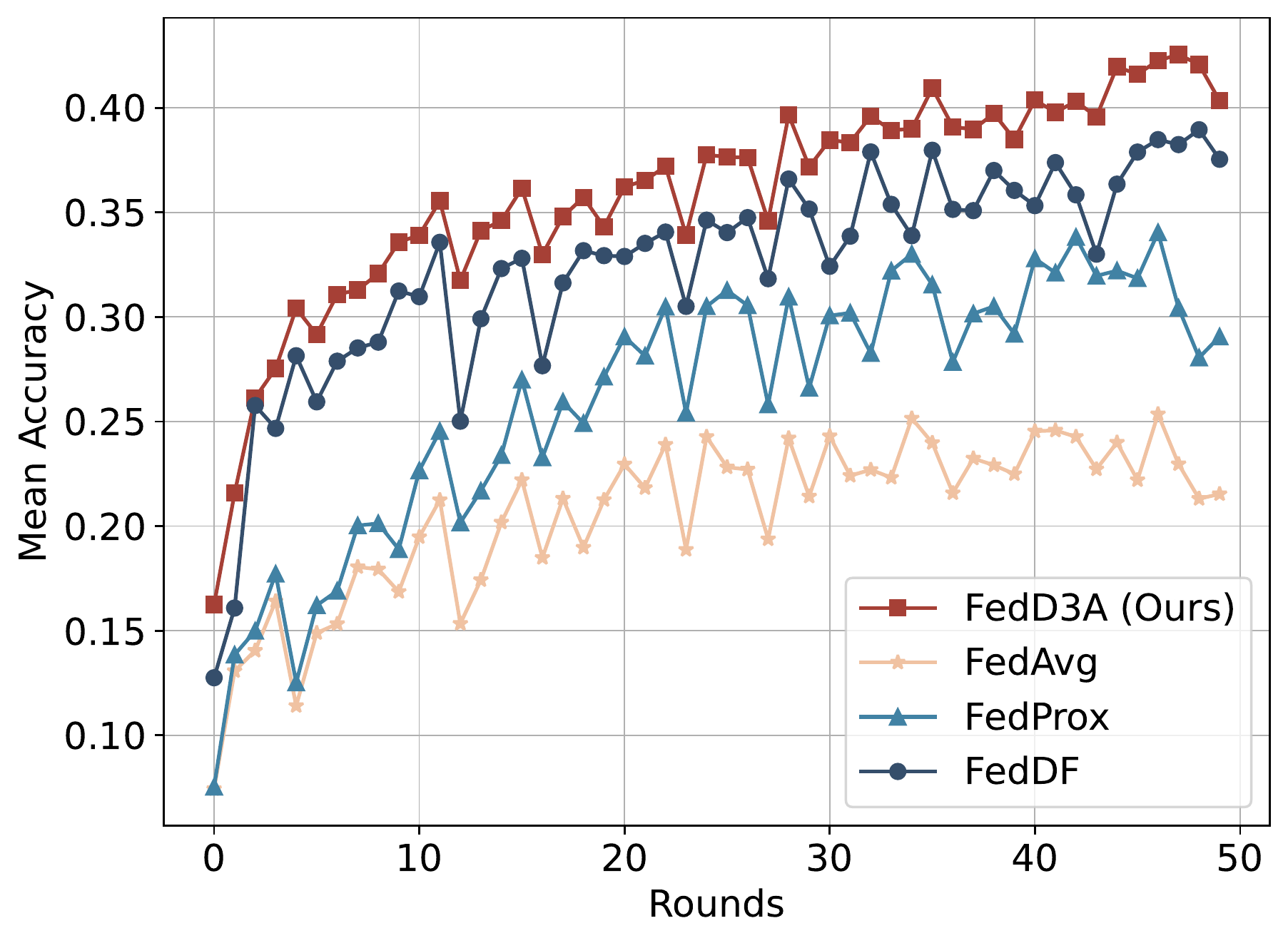}
}%
\subfigure[$\beta=1$]{
\includegraphics[width=0.24\linewidth]{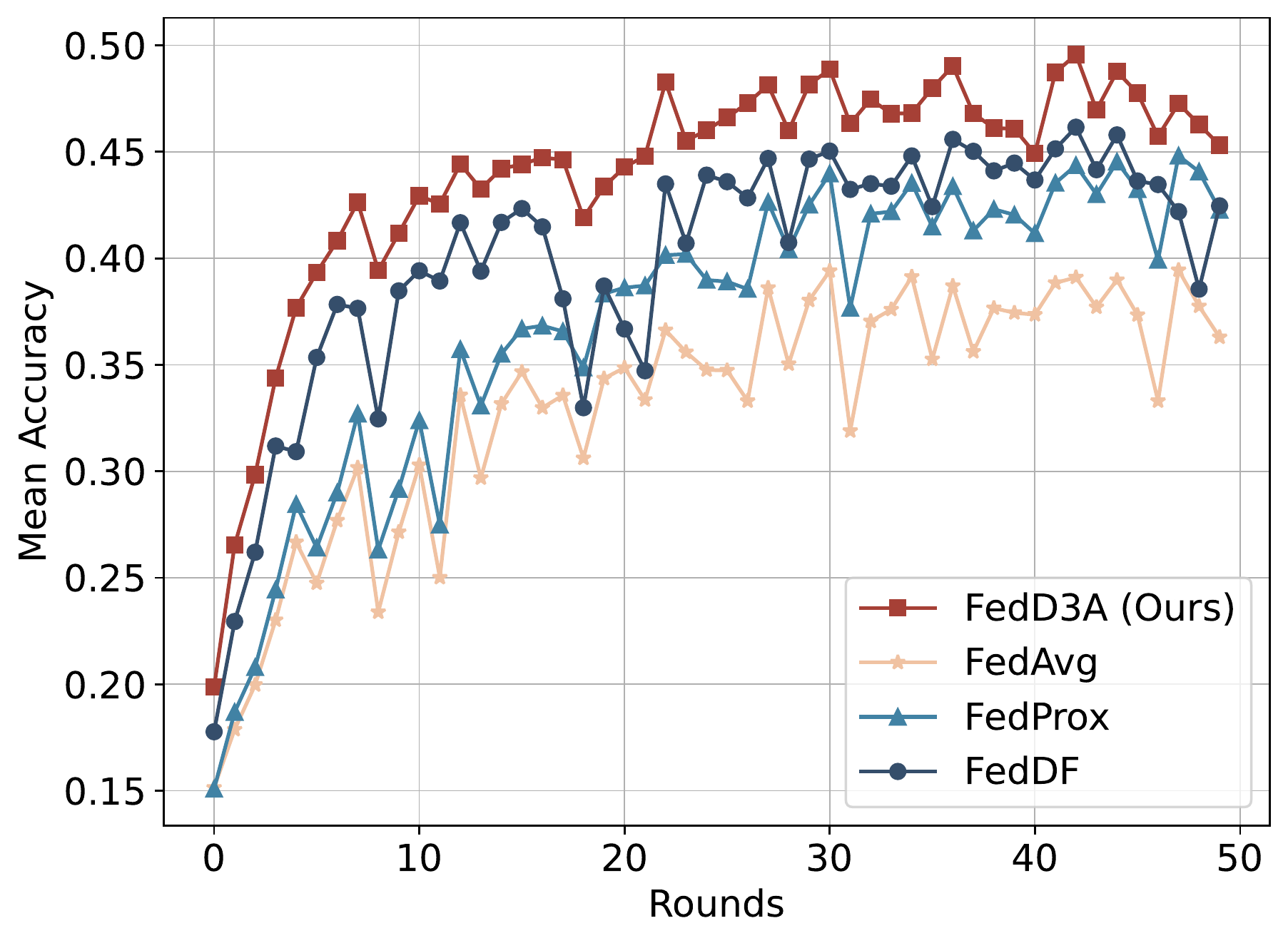}
}%
\subfigure[$\beta=0.2$]{
\includegraphics[width=0.24
\linewidth]{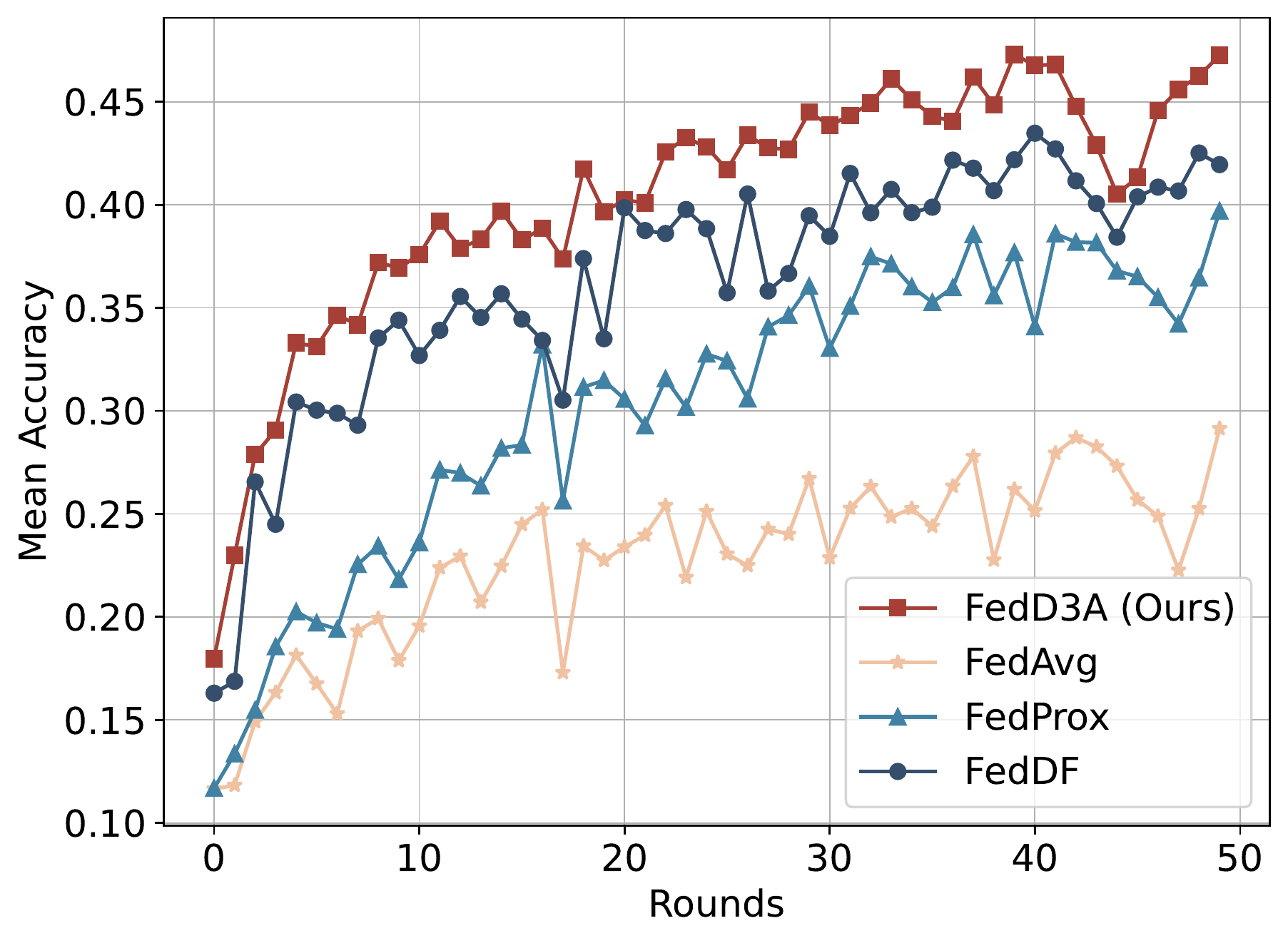}
}%
\subfigure[$\beta=1$]{
\includegraphics[width=0.24\linewidth]{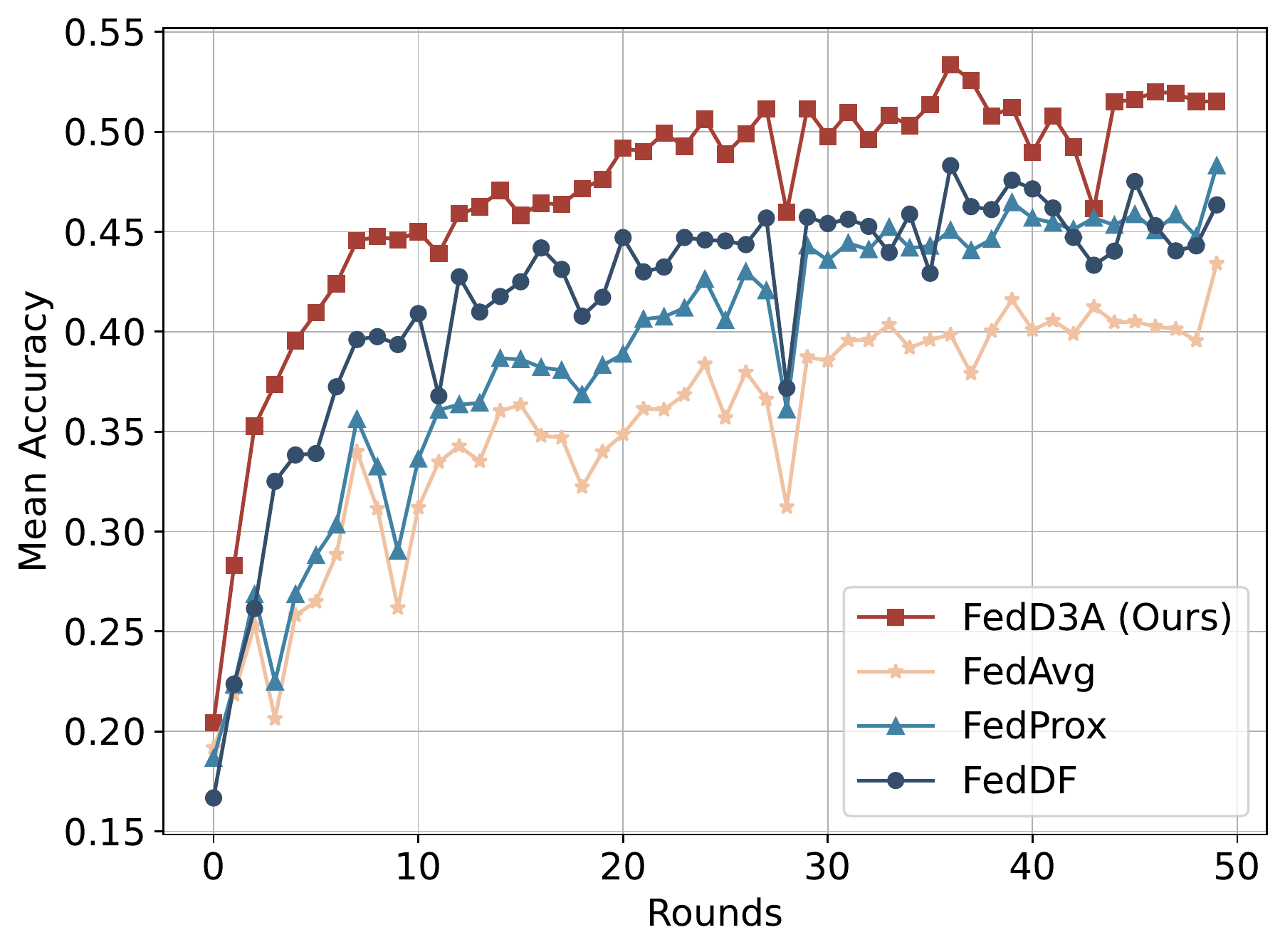}
}%
\caption{Average performance of the aggregated model on all clients in cross-device setting. (a) and (b) use \textit{real} as the server domain, (c) and (d) use \textit{clipart} as the server domain.}
\label{crossdevice}
\end{figure*}

\textbf{Implementation Details.} We take the momentum SGD as the optimizer. The initial learning rate is set to 0.01, the hyperparameter $\mu$ in FedProx is set to 0.1. With the increase of communication rounds, the learning rate gradually decreases to 1$e$-4 according to the cosine schedule. We implement our method based on PyTorch~\cite{paszke2019pytorch} with two NVIDIA GeForce RTX 3090 GPUs. The batch size is 256.

\textbf{Cross-silo Setting.} We select different domains as the server domain in turn, and the rest as the client domains. Take DomainNet for example, we can take the \textit{real} domain as the server, the remaining five domains serve as five clients. The number of local epochs and global rounds are 1 and 30, respectively. The distillation on the server also perform 1 epoch. The main results are shown in Figure~\ref{crosssilo}. Compared with the baseline methods, we can see: 1) In general, FedD3A has obvious improvement and can use fewer communication rounds to achieve higher accuracy, especially on the more challenging DomainNet dataset. 2) FedDF outperforms FedAvg. However, it is worse than FedProx. Also based on distillation, the accuracy of FedD3A can lead FedDF by nearly $10\%$ on DomainNet, much higher than FedProx, which indicates that addressing domain discrepancy problem is critical to improve the distillation.

%

\begin{figure}[tb]
\centering
\subfigure[FedD3A (ours)]{
\includegraphics[width=0.48\linewidth]{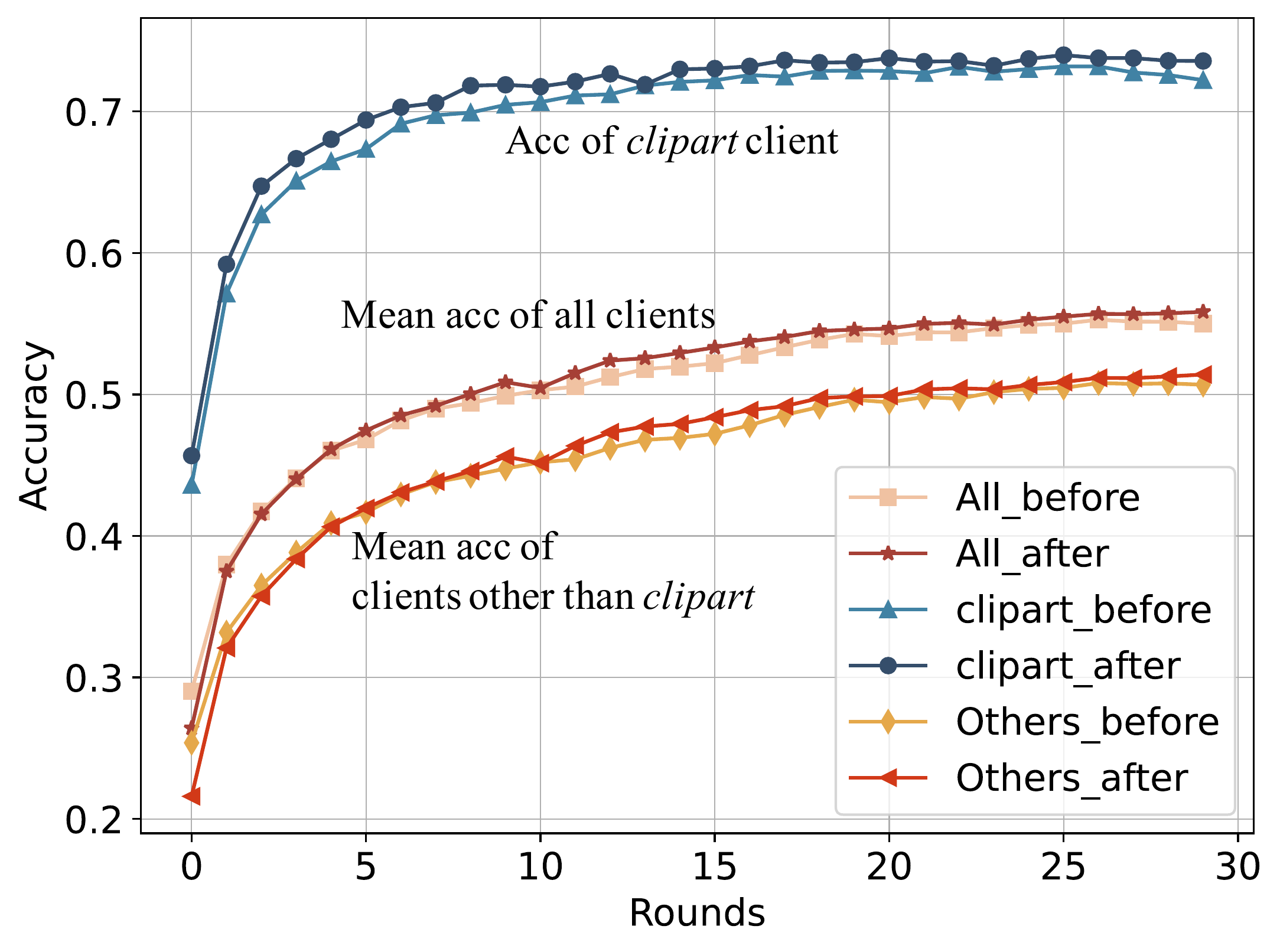}
}%
\subfigure[FedDF]{
\includegraphics[width=0.48\linewidth]{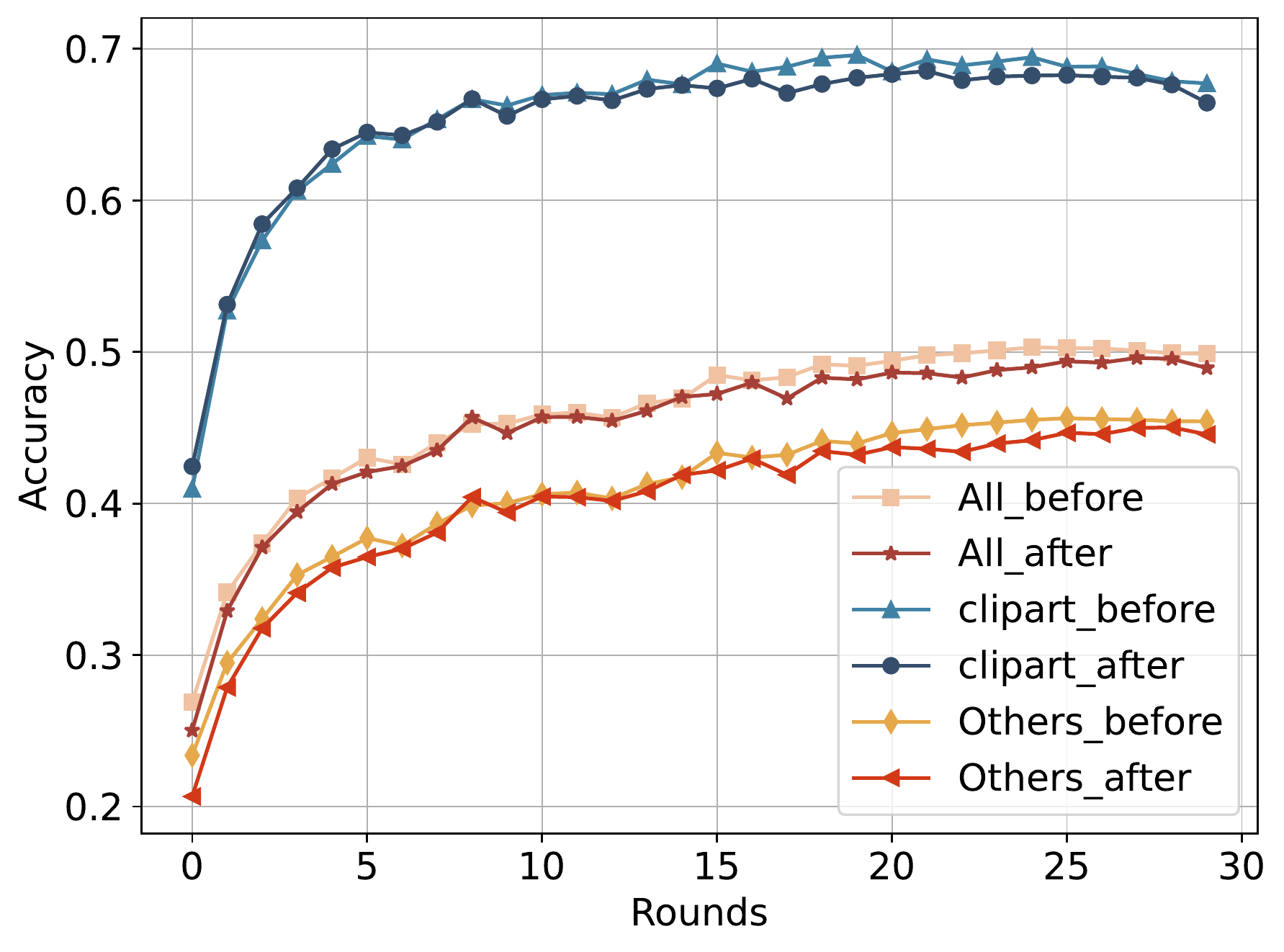}
}%
\caption{The results when the server data is completely consistent with the first client.  }
\label{repeatexp} 
\end{figure}

\begin{figure}[tb]
\centering
\subfigure[cross-silo]{
\label{heter1} 
\includegraphics[width=0.48\linewidth]{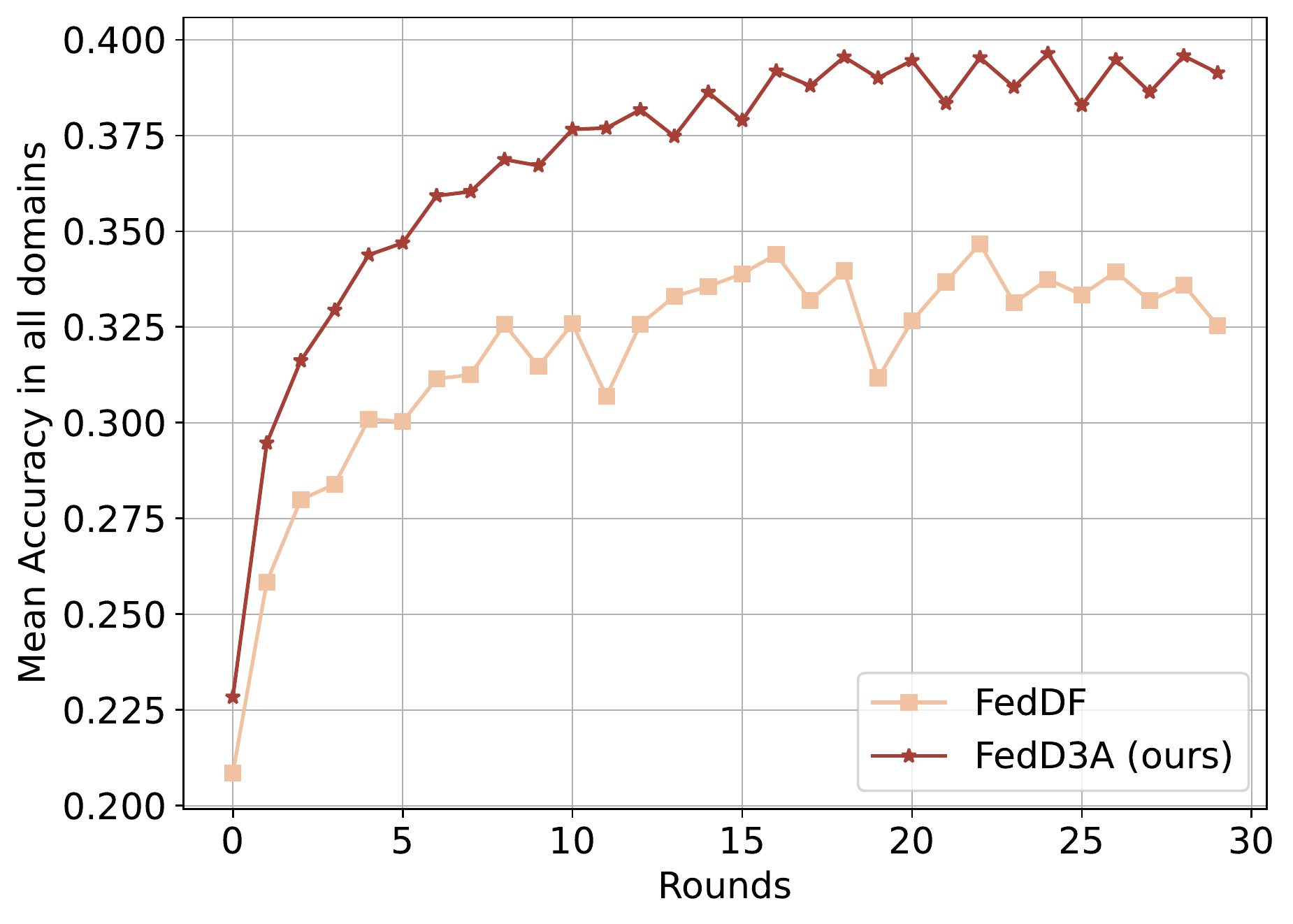}
}%
\subfigure[cross-device ($\beta=1$)]{ 
\includegraphics[width=0.48\linewidth]{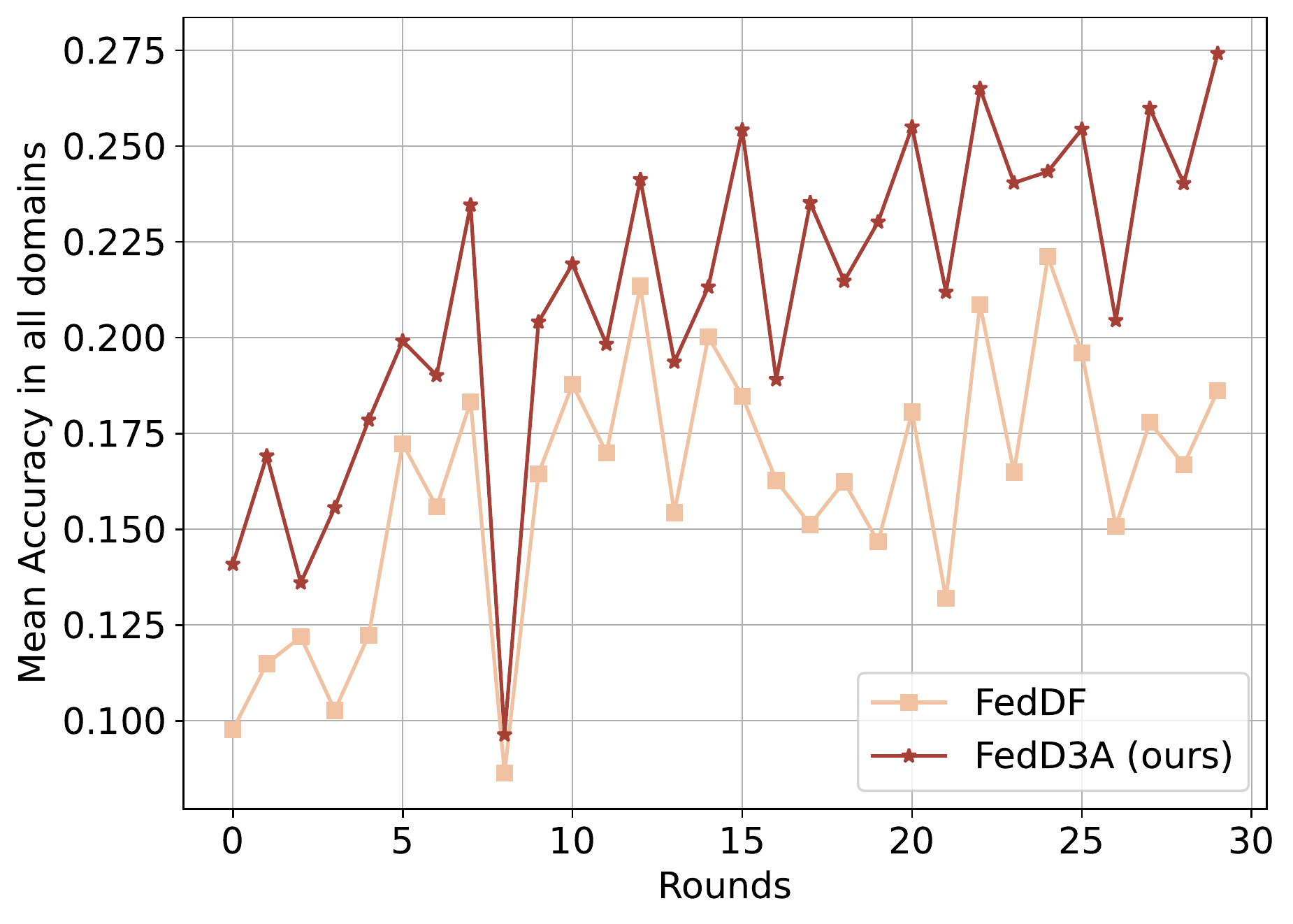}
}%
\caption{Heterogeneous model aggregation.  }
\label{heter} 
\end{figure}

In order to observe the performance of the aggregated model in extreme cases when the server domain is completely consistent with a client domain, we replace the data of the server in the above experiment Figure \ref{crosssilo}(a) with \textit{clipart}, that is, the data of the server and the first client are completely consistent. The result is shown in Figure~\ref{repeatexp}. `All\_before' denotes the mean accuracy on all clients before replacing the server data. It can be seen that: 1) In FedD3A, the mean accuracy of all clients is slightly improved compared with the original results and this improvement mainly comes from the \textit{clipart} domain. The performance of the other four domains has hardly changed. This is because the server data coincides with the first client (\textit{clipart}), so that the student can learn more from the teacher corresponding to that client. 2) In FedDF, all domains have a slight decrease.

\textbf{Cross-device Setting.} We split each local domain into five clients, then the DomainNet is divided into 25 clients. In order to simulate the non-IID case, following the common practices~\cite{lin2020ensemble}, we sample $\mathbf{p}_{c} \sim \operatorname{Dir}\left(\beta \mathbf{1}_{K}\right)$, $K=5$, and allocate a $p_{c,k}$ proportion of the instances with label $c$ to the training set of the $k$-th local model in each domain. When $\beta$ is small, the dataset is more challenging. In each round, 5 clients are selected for aggregation.

Figure~\ref{crossdevice} shows the results of $\beta=0.2$ and $\beta=1$. We can see: 1) Compared with $\beta = 1$, the gap between FedAvg and other methods is greater when $\beta = 0.2$. It can be seen that non-IID has a great impact on the performance of model averaging. 2) Compared with FedAvg, FedProx has a significant improvement, but its performance is still inferior to that of the distillation based method FedDF. While in the cross-silo experiment, FedDF is worse than FedProx. This indicates that the distillation based method is more effective when the clients are extreme non-IID.
3) In general, our FedD3A clearly outperforms other competitors with higher convergence speed and higher accuracy.

\textbf{Heterogeneous Model Aggregation.} In order to demonstrate the ability of FedD3A in heterogeneous model aggregation, we do experiments using VGG16~\cite{simonyan2014very}, ResNet34~\cite{he2016deep}, ResNet18~\cite{he2016deep} and MobileNetV3~\cite{howard2019searching} on different clients, respectively. For cross-device setting, five clients within each domain use the same kind of model. We take ResNet50 as the global model. In each round, the clients take the global model as a teacher, and then trains its own local model by cross-entropy loss and distillation loss (KL divergence). The server uses the heterogeneous models uploaded by the clients for multi-teacher distillation to obtain the global model. The result is in Figure~\ref{heter}. Compared with FedDF, FedD3A can converge to better performance and has a stable learning process. Only four rounds are needed to obtain the convergence accuracy of FedDF in Figure~\ref{heter1}.

 \begin{table}[t]
\centering
\caption{Comparison of different weighting strategies. }
\label{tab_aba}
\resizebox{\columnwidth}{!}{%
\begin{tabular}{lcccccc}
\Xhline{1pt}
\multicolumn{1}{c}{} & \begin{tabular}[c]{@{}c@{}}Client1\\ (\textit{infograph})\end{tabular} & \begin{tabular}[c]{@{}c@{}}Client2\\ (\textit{painting})\end{tabular} & \begin{tabular}[c]{@{}c@{}}Client3\\ (\textit{quickdraw})\end{tabular} & \begin{tabular}[c]{@{}c@{}}Client4\\ (\textit{real})\end{tabular} & \begin{tabular}[c]{@{}c@{}}Client5\\ (\textit{sketch})\end{tabular} & Avg \\ \hline
Ceiling & \textit{0.272} & \textit{0.544} & \textit{0.138} & \textit{0.707} & \textit{0.563} & \textit{0.445} \\\hline
Avg & 0.192 & 0.436 & 0.138 & 0.593 & 0.486 & 0.369 \\ 
Random & 0.194 & 0.443 & 0.126 & 0.583 & 0.468 & 0.363 \\ 
\rowcolor{gray}FedD3A-onehot & \textbf{0.264} & 0.501 & 0.162 & 0.634 & 0.540 & 0.420 \\ 
\rowcolor{gray}FedD3A-soft & 0.256 & \textbf{0.535} & \textbf{0.176} & \textbf{0.673} & \textbf{0.554} & \textbf{0.439} \\ 
 \Xhline{1pt}
\end{tabular}%
} 		 	 	 	  	 	 
\end{table}

\begin{table}[t]
\centering
\caption{Classification ability of subspace projection.}
\label{proto}
\resizebox{\columnwidth}{!}{%
\begin{tabular}{lcc}
\Xhline{1pt}
             & \multicolumn{1}{l}{prototype-based} & \multicolumn{1}{l}{projection-based} \\ \hline
MNIST (10 clients)        & $0.693_{\pm 0.005}$                            & $0.907_{\pm 0.004}  $                              \\
DomainNet  (345 clients)   &$0.382_{\pm 0.007}   $                             & $0.434_{\pm 0.009}   $                             \\
CIFAR10 (10 clients)      & $0.494_{\pm 0.063}   $                            & $0.628_{\pm 0.005} $                              \\
CIFAR100 (100 clients)     &$ 0.358_{\pm 0.007} $                             & $0.444_{\pm 0.003}$                                \\
Tiny Imagenet (200 clients) & $0.411_{\pm 0.024}    $                           & $0.504_{\pm 0.036}$                                \\
 \Xhline{1pt}
\end{tabular}
}
\end{table}

\textbf{Weighting Strategies.}  The main idea of FedD3A is adaptive filtering of teachers' knowledge. To see how much this contributes, we compare the following different weighting methods: 1) \textbf{Avg}. Take the average output of all teachers. 2) \textbf{Random}. Add random weights to the different teachers, i.e., $\boldsymbol{\alpha}=\text{SoftMax}(\mathbf{z})$, $\mathbf{z}$ is a random noise. 3) \textbf{FedD3A-onehot}. For each input sample, only the single teacher model with the largest weight is retained. 4) \textbf{FedD3A-soft}. Weight the different teachers using our proposed weighting method, see Eq.\ref{alpha}. This is used in the previous experiments. 5) \textbf{Ceiling}. Train a 5-class classifier by aggregating the datasets from 5 client domains. Note that this classifier is not actually available due to privacy issues. We do a one-round experiment (1 global round, 50 local epochs) using these five strategies. 

Take \textit{clipart} as server. The results are reported in Table~\ref{tab_aba}. It can be seen that random weighting is close to the average strategy, both of which have lower accuracy. In contrast, our FedD3A method has an obvious performance improvement. This shows that it is very important to carry out adaptive knowledge filtering for teacher models. In addition, FedD3A-soft is better than FedD3A-onehot, which implies that onehot strategy may ignore the overlapping knowledge among local models.


\textbf{Classification Ability of Subspace Projection.} In the above experiments, we verify the effectiveness of FedD3A in the FL context. One remaining problem is that, based on subspace projection, can we really judge which domain the test image comes from? To intuitively verify this, we design the following experiments. We divide the dataset according to category, and each client has one class. For example, the training set of CIFAR100 is divided into 100 clients according to category, and the client index is consistent with the class label. By uploading the subspace projection matrices, we try to classify the test set in the server \textbf{with only one round}. In this experiment, each category can be regarded as a domain, there is no need to consider cross-domain datasets. Therefore, we add several more common datasets in addition to Digit-5 and DomainNet. The backbone is a 3-layer MLP for MNIST, and a ResNet18 for CIFAR10/100, a ResNet50 for the other datasets. In this experiment, we take the prototype-based method as a baseline, in which, each client upload the mean feature as the class prototype. The server calculates which prototype the feature of the test image is most similar to. We take three random seeds and average the results. See Table~\ref{proto}, the projection-based method does not need to upload the raw features, and achieves better test accuracy.

\begin{figure}[tb]
\centering

\includegraphics[width=0.9\linewidth]{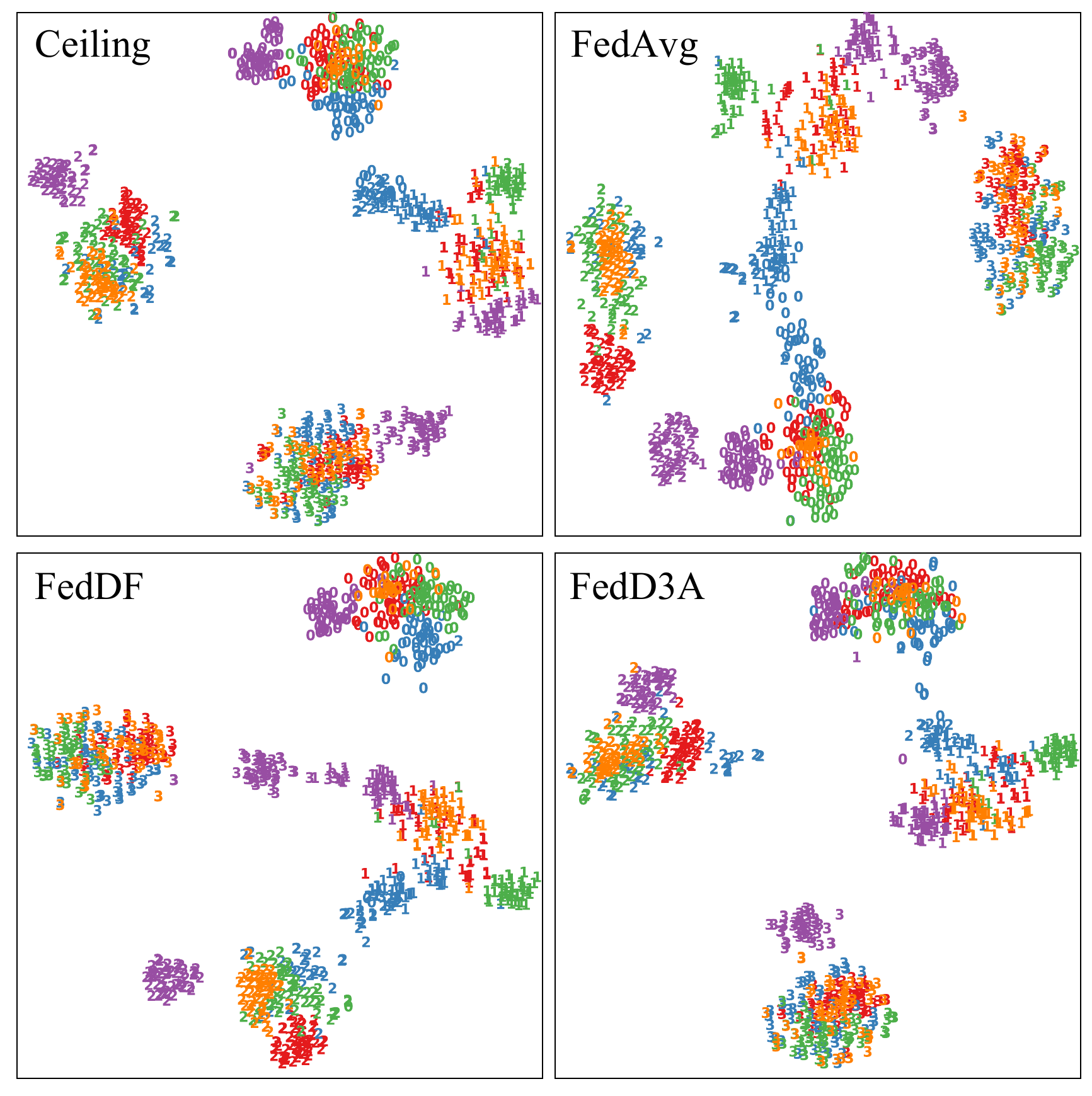}

\caption{Feature visualization of global model. Different numbers represent different classes. Different colors denote different domains. }
\label{vis}
\end{figure}  

\textbf{Visualization.} We randomly select four categories of data from all domains and visualize the features extracted by the global model in the cross-silo experiment.  The results are in Figure~\ref{vis}. `Ceiling' combines all client data to train a global model. It can be seen that: 1) Compared with the baseline method, FedD3A can learn better feature representations that are more conducive to classification. On the contrary, the features learned by FedAvg are greatly affected by domain discrepancy. 2) Domain information can be distinguished from the extracted features. In the same cluster, data from different domains are not mixed, which indicates that the subspace projection can be used to distinguish data domains.

\section{Conclusions}
In this paper, we explore the data domain discrepancy issue faced when leveraging server unlabeled data for model aggregation in federated learning. We show that the generalization error upper bound of the aggregation model on the clients can be reduced by assigning appropriate teacher weights to the unlabeled data on the server. Based on our analysis, we propose a domain discrepancy aware knowledge distillation algorithm FedD3A. Experimental results show that FedD3A outperforms the baseline methods in both cross-silo and cross-device FL settings, and heterogeneous model aggregation scenarios. Moreover, The experiments also verify that the adaptive filtering of teacher knowledge is very important to improve the performance of the aggregated model, which validate the claims in the contributions in Section~\ref{intro}. 

\bibliography{aaai23}

\end{document}